\documentclass[runningheads]{llncs}
\usepackage{graphicx}
\usepackage{booktabs}
\usepackage{hyperref}

\usepackage{amsmath}
\usepackage{amsfonts}
\begin{document}
\title{Interaction Matching for Long-Tail \\ Multi-Label Classification}

\author{
Sean MacAvaney\inst{1} \and
Franck Dernoncourt\inst{2} \and
Walter Chang\inst{2} \and \\
Nazli Goharian\inst{1} \and
Ophir Frieder\inst{1}}

\institute{IR Lab, Georgetown University, Washington DC 20057, USA \email{\{sean,nazli,ophir\}@ir.cs.georgetown.edu} \and
{Adobe Inc., San Jose, CA 95110, USA \\
\email{\{dernonco,wachang\}@adobe.com}} \\
}
\authorrunning{S. MacAvaney et al.}

\maketitle              
\begin{abstract}
We present an elegant and effective approach for addressing limitations in existing multi-label classification models by incorporating interaction matching, a concept shown to be useful for ad-hoc search result ranking. By performing soft n-gram interaction matching, we match labels with natural language descriptions (which are common to have in most multi-labeling tasks). Our approach can be used to enhance existing multi-label classification approaches, which are biased toward frequently-occurring labels. We evaluate our approach on two challenging tasks: automatic medical coding of clinical notes and automatic labeling of entities from software tutorial text. Our results show that our method can yield up to an 11\% relative improvement in macro performance, with most of the gains stemming labels that appear infrequently in the training set (i.e., the long tail of labels).
\end{abstract}

\section{Introduction}
Multi-label text classification (i.e., the task of assigning a variable number labels to a piece of text) is a classic task with a variety of practical applications. For instance, a clinical report could be tagged with medical codes, describing a patient's diagnoses (e.g., \textit{Lyme disease}) and procedures (e.g., \textit{clipping of aneurysm}). Since both the note and the medical codes are required in the clinical process, a system that can automatically derive these medical codes from notes would be a valuable time-saving measure. As another example, software tutorial text can be semantically labeled with the tools that are used to accomplish the task. These labels could provide additional support to users trying to replicate a tutorial, or be used to improve search engines by indexing on these labels.

Text labeling approaches rely on machine learning techniques to rank a given set of labels for a piece of text. For example, supervised FastText~\cite{joulin2017bag} learns dense label and document term representations. At inference time, it compares the label representations to the representation of an unseen document to produce label scores. Others have employed conceptually similar yet more sophisticated approaches, such as using convolutional neural networks and attention mechanisms in a similar fashion~\cite{Mullenbach2018ExplainablePO,Liu2017DeepLF}. One limitation of these approaches is that the models fail to effectively rank infrequently-occurring labels due to inadequate variability in the training data. These approaches are also unable to handle labels that do not occur in training data (e.g., extremely rare or new labels).

We present a text labeling approach that uses soft n-gram interaction matching, an approach inspired by recent work in ad-hoc ranking~\cite{Pang2016TextMA,Hui2018CoPACRRAC}. This allows handling of labels that have meaningful natural language names, while not necessarily occurring frequently in training data. It is common to have label names in multi-labeling tasks, as these are used by humans to manually perform the labeling (e.g., medical codes have descriptions). Our approach, which handles infrequent labels, can be combined with existing labeling techniques that handle frequently-occurring labels. We show that our approach is effective at two tasks, each with a large number of labels: automatic medical coding of clinical reports (1,159 labels), and automatic labeling of tools in software tutorials (831 labels).

In summary, our contributions are: (1) an approach for extending multi-label classification models, based on soft n-gram interaction matching; (2) an evaluation on two datasets, showing that our approach can be effectively combined with other leading classification approaches; and (3) an analysis demonstrating our capacity to identify long tail labels, even those without training samples.

\section{Background \& Related Work}

\textbf{Multi-label text classification.} This is a well-studied task with a multitude of prior work. Among the most notable recent efforts are supervised FastText~\cite{joulin2017bag}, which learns embeddings for labels that can be compared to document representations. Earlier work by Kim~\cite{Kim2014ConvolutionalNN} showed that a simple convolutional neural network (CNN) with dynamic pooling can be effective for text classification. Berger~\cite{Berger2015LargeSM} shows that recurrent neural networks (RNNs) can also be used for classification. Bow-CNN~\cite{Johnson2015EffectiveUO} uses n-gram indicator variables fed into a deep neural network to make classification decisions. PD-Sparse~\cite{Yen2016PDSparseA} addresses training data sparsity by enforcing heavy regularization penalties, but falls short of handling extremely infrequent labels. Liu et al~\cite{Liu2017DeepLF} attempts to address label sparsity by using shared intermediate representations from a CNN.
Gehrmann et al~\cite{gehrmann2018cnn} interpreted CNNs' classification by defining a saliency score.
Others have shown that using attention can further improve performance, and improve explainability of  decisions in the medical domain~\cite{Mullenbach2018ExplainablePO,Xie2018ANA}. Jain et al~\cite{Jain2019SliceSL} shows that millions of labels can be practically handled using a pruning strategy.
These approaches are limited by the variability of labels in the training data, or exact term matching.

\textbf{Soft n-gram interaction matching.} \textit{Interaction-focused ranking models} formulate document ranking as a learning-to-rank problem over a term similarity matrix between a query and a document. One successful approach to learn these patterns is by applying square convolutional kernels over the term similarity matrix (e.g., MatchPyramid\cite{Pang2016TextMA} and PACRR~\cite{Hui2018CoPACRRAC}), a process called \textit{soft-ngram interaction matching}. We propose an approach inspired by these methods for multi-label text classification; we use the approach to rank labels for a given segment of text, rather than documents for a query. Due to the large number of labels, we use a fixed sequence, rather than allowing the model to learn out-of-order n-grams. We also normalize the scores based on the label length.

\section{Methodology}
\label{sec:method}

\textbf{Notation.} Let $T$ be a sequence of tokens in a document, and let $\mathbf{L}$ be a collection of labels. For multi-label text classification, a score $V^{L_i,T}$ is generated for each label $L_i\in \textbf{L}$. The labels are ranked by this score, optionally employing a score cutoff threshold and/or a maximum count to select the labels.

For our task, we assume each label consists of a sequence of tokens representing its name in natural language: $L_i=\{L_{i,1},L_{i,2},...,L_{i,|L_i|}\}$. This is a reasonable assumption, given that the labels are usually produced for humans, who will often need a name to reason about the label. For our method to be most effective, the names should have terms that may have approximate matches in the text. For instance, given the procedure label of \textit{clipping of aneurysm}, an approximate match found in the text might be \textit{clip the aneurysm}. Our method uses the term similarity matrix $S^{L_i,T}\in \mathbb{R}^{|L_i|,|T|}$ between label $L_i$ and text $T$ as input:
\begin{equation}\small
S^{L_i,T}_{j,k}=\cos(embed(L_{i,j}),embed(T_k))
\end{equation}
where $\cos(\cdot)$ is the cosine similarity score and $embed(\cdot)$ returns the token's word embedding. Note that each similarity score here is a unigram match; our model operates over this matrix to perform n-gram matching.

\textbf{Interaction matching model.} Inspired by  recent interaction-focused ranking models in information retrieval, we apply square convolutions over the similarity matrix to produce soft n-gram matching scores. Unlike document ranking models, however, we impose a single fixed sequential convolution kernel over the labels. In other words, we use the identity matrix $I_{|L_i|}$ as a convolutional kernel. We then take the maximum scores from each kernel and normalize them by the length of the label $|L_i|$. This step is not taken in the document ranking models but necessary in this context because multiple labels of different lengths are being matched over the same document. Note that since the convolutional kernel matches similarity scores, exact matches are not required; this is what makes the n-grams `soft'. More formally, our method generates a label score for each document position $P^{L_i,T}\in \mathbb{R}^{|T|}$ for document $T$ and label $L_i$:
\begin{equation}\small
P^{L_i,T}=\sigma\Big(\frac{I_{|L_i|} \star S^{L_i,T}}{|L_i|}\Big)
\end{equation}
where $\sigma(\cdot)$ is the sigmoid activation function and~$\star$ performs 2-dimensional convolution. For simplicity, we assume~$\star$ applies padding where necessary.

To generate a label score for the entire text, we perform max pooling:
$V^{L_i,T}=\max_{j=1}^{|T|}P^{L_i,T}_j$.
The use of max pooling allows for the soft n-gram to match anywhere in the document and for the score not to be influenced by document length (as opposed to average pooling, for instance). Furthermore, the $\arg\max$ yields an interpretable grounding of the model's decision within the text and can be used to aid in the explanation of the model's decision. At inference time, all labels in $\textbf{L}$ are ranked using this method. This approach is trivially parallelizable and easily handles datasets with thousands of labels.

The model's structure allows the interaction model to easily incorporate new labels to be introduced after training simply by adding to $\textbf{L}$. Although the model has no learnable parameters on its own, in our experiments, we train the model by back-propagating errors to the word embeddings, producing representations that are enhanced for this task. We recognize that our model is ineffective for labels that do not match the text. Thus, we suggest incorporating our method into existing multi-label text classification approaches, which can learn to effectively match labels that frequently occur in the training data. We train the two models jointly, combining them by taking the maximum score for each label.

\section{Experiments}
\label{sec:exp}

We test our approach on two tasks: medical coding and software tool extraction from online tutorials. While both are multi-label classification tasks, the data characteristics are different for each task, demonstrating that our approach is generally applicable. Examples from both datasets are given in Figure~\ref{fig:mimic-ex}.

\begin{figure}
\scriptsize
\begin{tabular}{p{2.4in}p{2.1in}}
\multicolumn{1}{c}{\bf Medical Coding} & \multicolumn{1}{c}{\bf Software Tutorial} \\
\cmidrule(lr){1--1}\cmidrule(lr){2--2}
\textbf{Report:}
...status post tracheostomy for paradoxical vocal cord motion with asthma discharge medications
fenofexadine mg po q day calcium carbonate grams po t i d percocet one po q to hours prn pain... &
\textbf{Tutorial:}
Create another new document (I chose 600 x 400 px for width and height), select the brush tool, and
open the brush preset panel. \\
\textbf{ICD-9 labels:} Other diseases of upper respiratory tract ; Asthma ; Diabetes mellitus &
\textbf{Tool labels:} File $>$ New ; Brush Tool \\
\cmidrule(lr){1--1}\cmidrule(lr){2--2}
\end{tabular}
\centering\scriptsize

\caption{Example de-identified clinical report excerpt from MIMIC-III, including top-level ICD-9 labels, and sentence from the tutorial dataset with labeled software tools.}
\label{fig:mimic-ex}
\end{figure}

\subsection{Medical coding}
\textbf{Dataset.} We first evaluate on the MIMIC-III dataset~\cite{Johnson2016MIMICIIIAF}, a large, de-identified, and publicly-available\footnote{Available with a data usage agreement, https://mimic.physionet.org/} collection of medical records. Each record in the dataset includes ICD-9 codes,\footnote{https://www.cdc.gov/nchs/icd/icd9.htm} which identify diagnoses and procedures performed. Each code is partitioned into sub-codes, which often include specific circumstantial details. We treat the parent (top-level) codes as labels to be identified based on the patient's discharge note.
The dataset consists of 112k clinical reports records (avg length 709.3 tokens) and 1,159 top-level ICD-9 codes (labels). Each report is assigned to 7.6 codes, on average. See Figure~\ref{fig:mimic-ex} for an excerpt of a report with labeled codes. We use the same train/dev/test split used by~\cite{Mullenbach2018ExplainablePO}, with 1,632 development and 3,372 testing reports. We train word embeddings on MIMIC-III using word2vec~\cite{Mikolov2013DistributedRO}, matching the setting of~\cite{Mullenbach2018ExplainablePO}. Note that this does not preclude the matching of terms unseen in training data; trivially, a larger unlabeled corpus could be employed for training embeddings or binary matching could be used for out-of-vocabulary terms.

\textbf{Baselines \& training.} We compare our approach to the state-of-the-art attention-based CAML~\cite{Mullenbach2018ExplainablePO} network for medical coding, along with a convolutional neural network text classifier network~\cite{Kim2014ConvolutionalNN} and a bi-directional GRU network as baselines, using the implementations provided by~\cite{Mullenbach2018ExplainablePO}. These represent strong baselines for this task. We combine our approach with each baseline, training them jointly as described in Section~\ref{sec:method}. We train all neural models optimizing cross entropy loss with the Adam optimizer~\cite{Kingma2015AdamAM} (learning rate of $10^{-4}$). We select a threshold for all labels using MacroF1 performance on the dev set.

\begin{table}
\caption{Medical coding performance on MIMIC-III. Our results on the macro metrics for CNN and CAML are significant at $p<0.05$.}
\label{tab:mimic}
\centering
\begin{tabular}{lrrrr}
\toprule
Model & MacroP & MacroR & MacroF1 & MicroF1 \\
\midrule
Bi-RNN & 0.1585 & 0.1486 & 0.1534 & 0.5890 \\
+ interaction (ours) & 0.1831 & 0.1233 & 0.1473 & 0.5970 \\
\midrule
CNN & 0.1628 & 0.1562 & 0.1594 & 0.5894 \\
+ interaction (ours) & 0.1822 & 0.1745 & 0.1783 & 0.6072\\
\midrule
CAML & 0.2582 & 0.2235 & 0.2396 & 0.6520 \\
+ interaction (ours) & \bf 0.2720 & \bf 0.2310 & \bf 0.2498 & \bf 0.6546 \\
\bottomrule
\end{tabular}

\end{table}

\textbf{Results.} Our results on MIMIC-III are presented in Table~\ref{tab:mimic}. In line with prior work on the dataset, we measure the performance in terms of micro- and macro-averaged F1 score. Since our focus is improving the long tail of infrequently-occurring labels, we also include macro-averaged precision and recall. Our approach outperforms the state-of-the-art CAML method in terms of macro precision, recall and F1 by 3--5\% (relative improvement, significant at $p<0.05$). The performance improvement on the weaker CNN baseline is even more pronounced, achieving an 8--11\% improvement on the macro metrics (also significant at $p<0.05$). Interestingly, our approach improves the precision for the bi-directional RNN, at the expense of recall. We attribute this to the interaction matching technique that is inherently high-precision. The improvements in the micro metrics are less pronounced (and worse, in the case of micro recall on CAML), showing that our approach primarily benefits the long tail.

\textbf{Error analysis.} We often found that our method was able to match infrequent labels where CAML had failed. For instance, in one report, our method labeled all three codes correctly (including one that occurs in only 0.5\% of training data), while the unmodified CAML method found two of the three correctly, but also mistakenly included a third, completely unrelated label (occurs in about 0.1\% of training data). We observed cases where general codes were not matched effectively by either model. For instance, \textit{Other diseases of lung} is difficult to match by both models because it involves more advanced reasoning (i.e., the condition affects the lung, and there isn't another label).

\subsection{Software tutorial labeling}
\textbf{Dataset.} We also evaluate our approach on a collection of software tutorials, labeled with the tools used to complete each step (by sentence). This dataset is collected from online tutorials and manually labeled by sentence with a large collection of software tools (831 in total). The dataset consists of 40k sentences, with an average length of 21.7 tokens and an average number of 1.2 tools labeled per record. An example labeled tutorial sentence is given in Figure~\ref{fig:mimic-ex}. Note that the tool mentions can be either explicit (\textit{brush tool} $\rightarrow$ \texttt{Brush Tool}) or implicit (\textit{Create a new document} $\rightarrow$ \texttt{File > New}). The dataset will be made available for validation of our results. We use a random 90/5/5\% train/dev/test set split.

\textbf{Baseline.} Since no specialized systems exist for this dataset, we use supervised FastText~\cite{joulin2017bag} and XML-CNN~\cite{Liu2017DeepLF} as a baseline approaches for the tutorial dataset. FastText is trained for 100 epochs with 1--3 word n-grams, and XML-CNN is trained using default settings. We fine-tune the word embeddings for the interaction approach by taking the maximum label score between the unmodified model score and the score from the interaction model for each label.

\begin{table}
\centering\small
\caption{Average precision (macro-averaged by label) performance on the Tutorial dataset. Significant results that are significant at $p<0.01$ are indicated by $\uparrow$.}
\label{tab:tutorial-results}
\begin{tabular}{lrr}
\toprule
Model & Dev Set & Test Set \\
\midrule
FastText & 0.5569 & 0.5444 \\
+ interaction (ours) &$\uparrow$ \textbf{0.6225} &$\uparrow$ \textbf{0.5789} \\
\midrule
XML-CNN & 0.5882 & 0.5916 \\
+ interaction (ours) &\bf0.5910 &\bf0.5923 \\
\bottomrule
\end{tabular}
\end{table}

\textbf{Results.} We present our results on the tutorial dataset in Table~\ref{tab:tutorial-results}. 
We use average precision (AP, macro-averaged by label), given the environment that users are presented with a list of possible software tools to complete the step of the tutorial. When applied to FastText, our approach improves the test set performance by 6\% (relative improvement, significant at $p<0.01$).
The FastText model achieved a perfect AP score of 1.0 for 55 of the labels found in the test set (meaning it was ranked highest among all labels whenever it appeared), whereas the interaction variant had a perfect score for 69 labels. This is a 25\% improvement, most of which came from the less frequent half of the labels. The least frequent quarter saw an even bigger change, from 11 perfect scores to 26 (136\%), four of which had no training samples. Our approach also slightly improves the performance on XML-CNN, though the results are not significant at $p<0.01$. Interestingly, the XML-CNN model appears to hamper performance in the development set.

\section{Conclusion}
We presented an approach to enhance existing multi-label classification techniques that employs soft n-gram interaction matching. We demonstrated that the approach is effective at identifying labels in the long tail, which are under-represented with current state-of-the-art classification approaches. We also showed that the approach can effectively label items that do not appear at all in the training data.

\bibliographystyle{splncs04}
\bibliography{biblio}

\begin{thebibliography}{10}
\providecommand{\url}[1]{\texttt{#1}}
\providecommand{\urlprefix}{URL }
\providecommand{\doi}[1]{https://doi.org/#1}

\bibitem{Berger2015LargeSM}
Berger, M.J.: Large scale multi-label text classification with semantic word
  vectors (2015)

\bibitem{gehrmann2018cnn}
Gehrmann, S., Dernoncourt, F., Li, Y., Carlson, E.T., Wu, J.T., Welt, J.,
  Foote~Jr, J., Moseley, E.T., Grant, D.W., Tyler, P.D., et~al.: Comparing deep
  learning and concept extraction based methods for patient phenotyping from
  clinical narratives. PloS one  \textbf{13}(2),  e0192360 (2018)

\bibitem{Hui2018CoPACRRAC}
Hui, K., Yates, A., Berberich, K., de~Melo, G.: Co-pacrr: A context-aware
  neural ir model for ad-hoc retrieval. In: WSDM (2018)

\bibitem{Jain2019SliceSL}
Jain, H., Balasubramanian, V., Chunduri, B.R., Varma, M.: Slice: Scalable
  linear extreme classifiers trained on 100 million labels for related
  searches. In: WSDM (2019)

\bibitem{Johnson2016MIMICIIIAF}
Johnson, A.E.W., Pollard, T.J., Shen, L., wei H.~Lehman, L., Feng, M.,
  Ghassemi, M.M., Moody, B., Szolovits, P., Celi, L.A., Mark, R.G.: Mimic-iii,
  a freely accessible critical care database. In: Scientific data (2016)

\bibitem{Johnson2015EffectiveUO}
Johnson, R., Zhang, T.: Effective use of word order for text categorization
  with convolutional neural networks. In: HLT-NAACL (2015)

\bibitem{joulin2017bag}
Joulin, A., Grave, E., Bojanowski, P., Mikolov, T.: Bag of tricks for efficient
  text classification. In: EACL (2017)

\bibitem{Kim2014ConvolutionalNN}
Kim, Y.: Convolutional neural networks for sentence classification. In: EMNLP
  (2014)

\bibitem{Kingma2015AdamAM}
Kingma, D.P., Ba, J.: Adam: A method for stochastic optimization. In: ICLR
  (2015)

\bibitem{Liu2017DeepLF}
Liu, J., Chang, W.C., Wu, Y., Yang, Y.: Deep learning for extreme multi-label
  text classification. In: SIGIR (2017)

\bibitem{Mikolov2013DistributedRO}
Mikolov, T., Sutskever, I., Chen, K., Corrado, G.S., Dean, J.: Distributed
  representations of words and phrases and their compositionality. In: NIPS
  (2013)

\bibitem{Mullenbach2018ExplainablePO}
Mullenbach, J., Wiegreffe, S., Duke, J., Sun, J., Eisenstein, J.: Explainable
  prediction of medical codes from clinical text. In: NAACL-HLT (2018)

\bibitem{Pang2016TextMA}
Pang, L., Lan, Y., Guo, J., Xu, J., Wan, S., Cheng, X.: Text matching as image
  recognition. In: AAAI (2016)

\bibitem{Xie2018ANA}
Xie, P., Shi, H., Zhang, M., Xing, E.P.: A neural architecture for automated
  icd coding. In: ACL (2018)

\bibitem{Yen2016PDSparseA}
Yen, I.E.H., Huang, X., Ravikumar, P., Zhong, K., Dhillon, I.S.: Pd-sparse : A
  primal and dual sparse approach to extreme multiclass and multilabel
  classification. In: ICML (2016)

\end{thebibliography}

\end{document}